\documentclass[letterpaper, conference]{ieeeconf}
\IEEEoverridecommandlockouts
\usepackage{cite}

\usepackage{easyReview}
\usepackage{amsmath,amssymb,amsfonts}
\usepackage{algorithmic}
\usepackage{graphicx}
\usepackage{textcomp}
\usepackage{xcolor}

\begin{document}

\title{\LARGE \bf Deformable Tip Mount for Soft Growing Eversion Robots
}
\author{Cem Suulker,  Sophie Skach, Danyaal Kaleel, Taqi Abrar, Zain Murtaza, \\ Dilara Suulker, and Kaspar Althoefer, \it{Senior Member, IEEE}%
\thanks{This work was supported in part by the EPSRC National Centre for Nuclear Robotics project (EP/R02572X/1), the Innovate UK GrowBot project (96566) and the Innovate UK LianaBot project (89911).}
\thanks{Authors are with the Centre for Advanced Robotics @ Queen Mary, School of Engineering and Materials Science, Queen Mary University of London, United Kingdom.
        {\tt\footnotesize c.suulker@qmul.ac.uk}}%
}
\maketitle

\begin{abstract}

Here we present a flexible tip mount for eversion (vine) robots. This soft cap allows attaching a payload to an eversion robot while allowing moving through narrow openings, as well as the eversion of protruding objects, and expanded surfaces.

\end{abstract}

\section{Introduction}

Growing robots based on the eversion principle are known for being capable of immensely extending along their longitudinal axis and reaching far into inaccessible, remote spaces. 

One major challenge has been to attach a meaningful payload, such as sensors, at the tip which is never steady but rather evolving because of the way an eversion robot is constantly moving its inside to the outside to progress forward. Various types of mechanisms have been proposed by the robotics community, caps that engulf the tip section of the eversion robot and are often linked to the everting material of the eversion robot via complex sets of rollers or magnets. As a disadvantage, these mechanisms do not work reliably, especially when the eversion robot is integrated with functional elements that protrude from the robot’s skin such as navigation pouches, sensors or electronics. Another shortcoming of existing caps is that they are made from rigid materials, preventing the robot from moving through narrow openings and reducing its navigation capabilities. 

We present here the soft cap for eversion robots. Our soft cap made from fabric and with a diameter similar to that of the eversion robot is easily slipped over the tip like a beanie. Our experimental study shows that the soft cap sits on the eversion robot in a robust fashion and is capable of transporting a payload (here: a camera) across long distances. We also show that the robot’s capability to move through narrow openings or to bend its body is virtually unhindered by the slipped-on cap. 

\begin{figure}[t]
  \centering
  \includegraphics[width=1\linewidth]{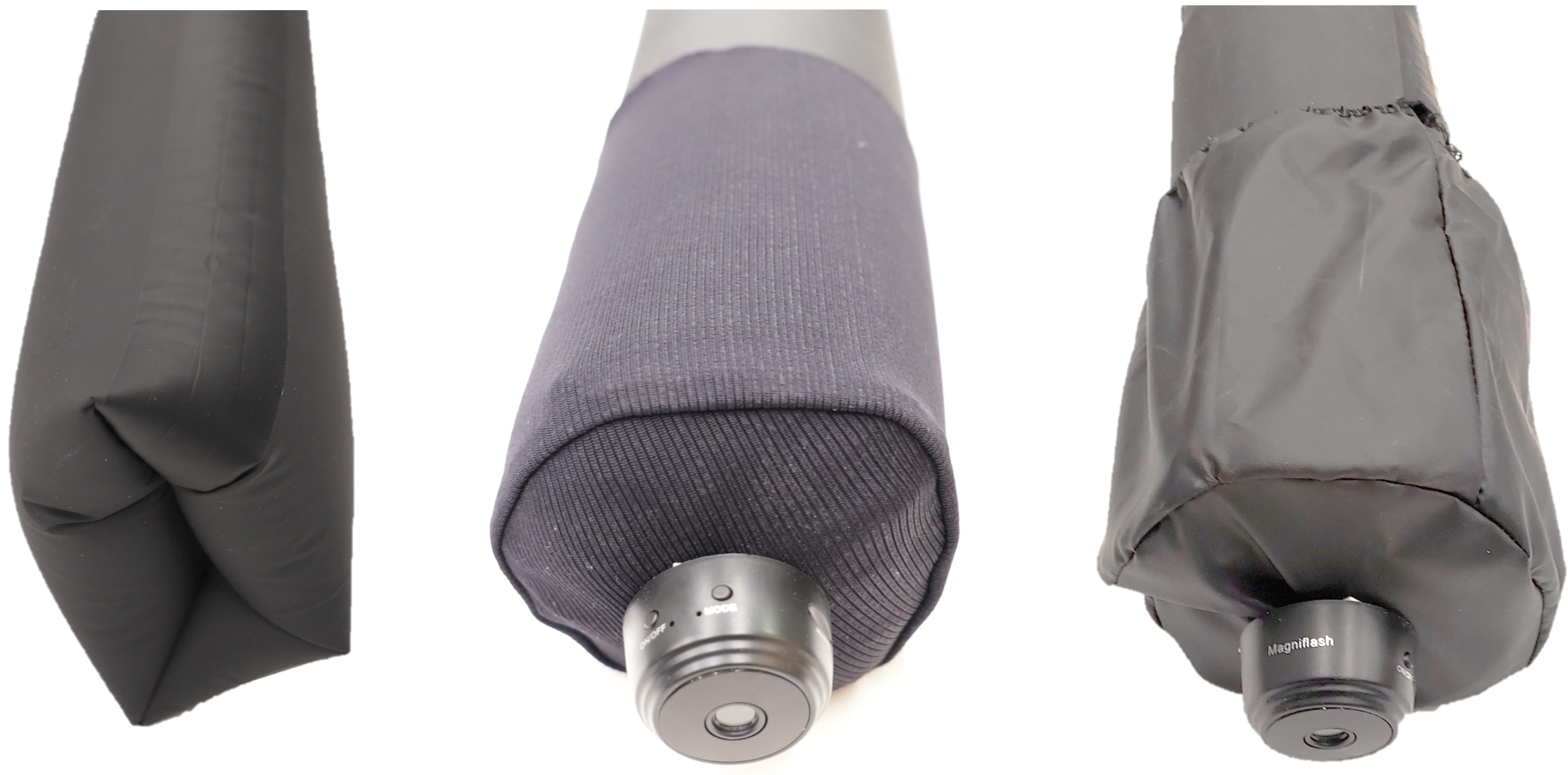}
  \caption{Left, an inflated eversion robot. Middle, a soft cap made from flexible fabric slipped over the tip of eversion robot. Right, a soft cap made using non-flexible fabric and elastic bands, slipped over the tip of eversion robot. A tool, here a camera, is attached to the caps.}
  \label{fig1}
\end{figure}

\section{Soft Cap Design}

Here, we take on the challenge to develop such soft cap and design around the shortcomings of its rigid counterparts, taking advantage of the given structural properties of an increasingly popular material for soft robotics: textiles.

We exploit the friction force between the cap and the robot body as a way to keep the cap in place without restricting robot movement. With the help of elastic fabrics \cite{suulker}, the cap is slipped onto the tip of the everting robot, like a beanie, able to slightly adapt its diameter through its stretch character. This encapsulation of the body is still firm, yet flexible.

\begin{figure}[h]
  \centering
  \includegraphics[width=0.63\linewidth]{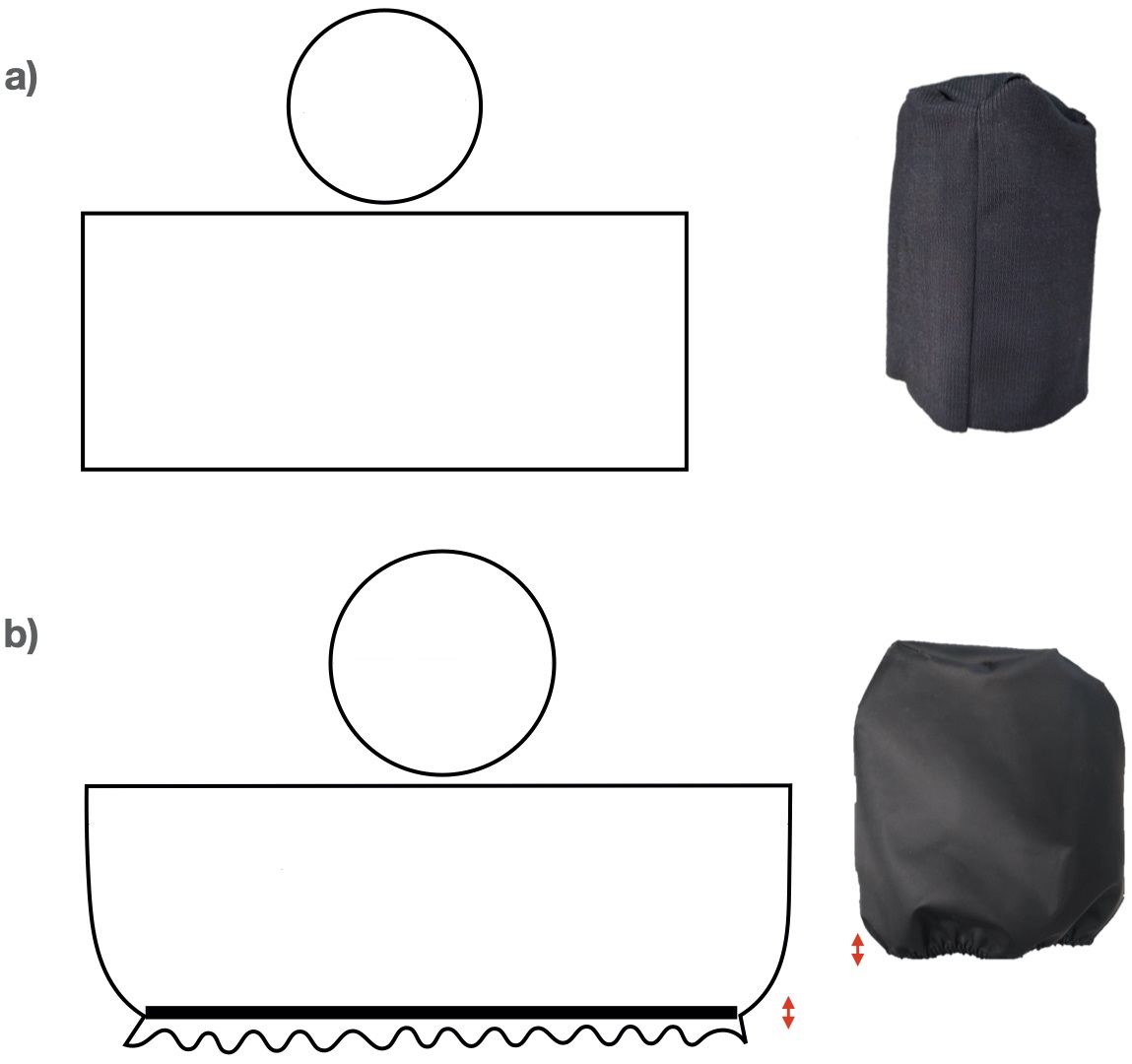}
  \caption{Pattern constructions a) using elastic finely knitted rib fabric material, and b) using non-elastic woven fabric and elastic band.}
  \label{figCP}
\end{figure}

There are various cutting and sewing patterns that could be explored, this extended abstract focuses on the patterns shown in Fig. \ref{figCP}. 
In the first design (Fig. \ref{figCP} a) one-way elastic fabric is used to expand the cap diameter. Second design (Fig. \ref{figCP} b) the cap is made from non-stretch fabric but by using elastic bands, the cap's bottom section will squeeze onto the eversion robot body. 
An elastic band is commonly used in clothing to create ruffles or elastic waistbands. Being a standard method in the textile industry, it has rarely been employed for the development of textile robotics \cite{suulkerral}. Using this method leads to a smaller effective contact surface. The thickness of the elastic band gives the effective contact length (Red arrows in Fig. \ref{figCP} b).

\section{Case Study \& Results}

In this section, we showcase the effectiveness of our soft caps by subjecting them to challenging scenarios that would typically pose difficulties for rigid caps. These challenges were designed to closely resemble realistic scenarios.

\subsection{Quality 1: Eversion of protruding solid objects}

This scenario simulates the placement of sensors, electrical components, air tubes, and rigid connectors on the body of the eversion robot. These rigid components can cause jamming in caps with complex mechanisms,  entirely blocking the movement of the robot.

\begin{figure}[h]
  \centering
  \includegraphics[width=0.8\linewidth]{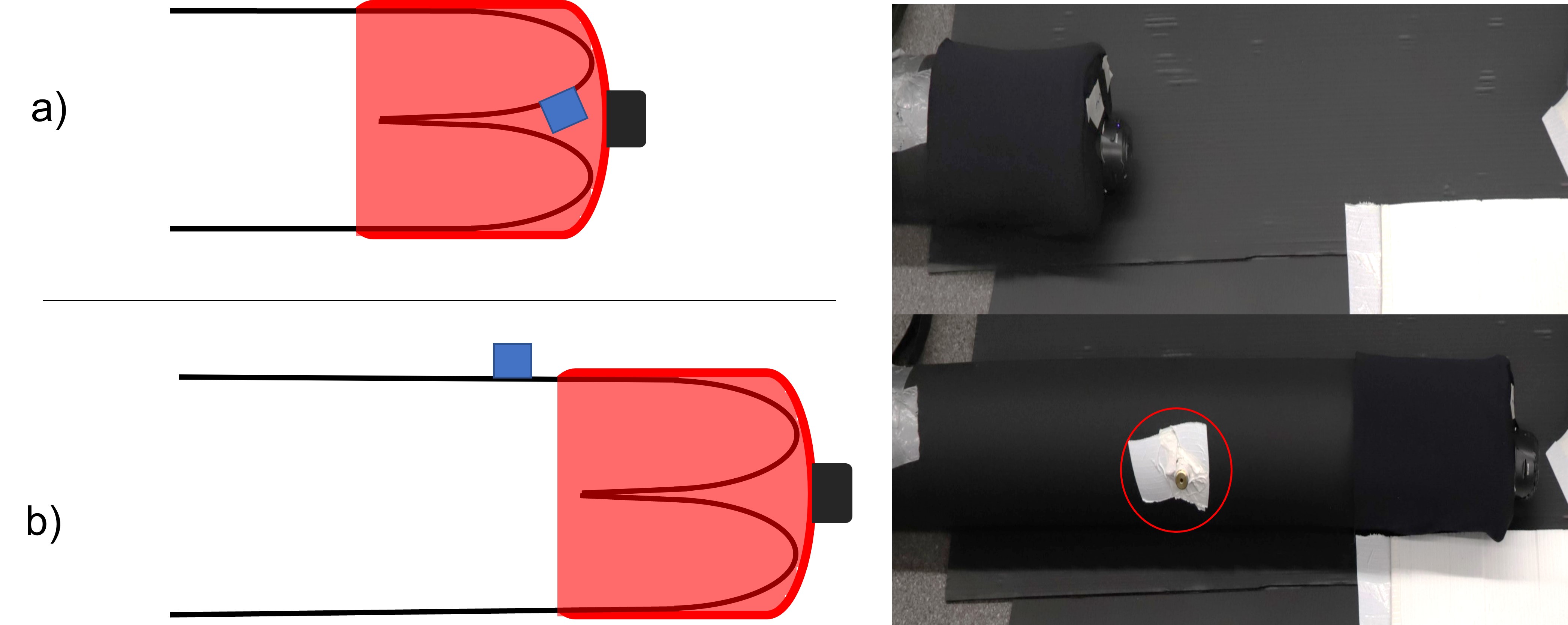}
  \caption{a) An eversion robot with a soft cap with a camera attached to it b) The cap successfully allows eversion of 1.7 cm long protruding material.}
  \label{figCh2}
\end{figure}

To simulate this scenario, we roughly taped two 1 cm wide, 1.3 cm long and 1.6 cm wide, 1.7 cm long pipe connectors to the robot body in random places.
In the starting condition, the robot is inflated and protruding solid objects are not everted yet (Fig. \ref{figCh2} a). The robot with a cap grows despite the protruding objects, and the objects should be fully visible leaving the boundaries of the cap (Fig. \ref{figCh2} b).

\subsection{Quality 2: Squeezability}

Eversion robots are often marketed for their ability to easily navigate through narrow openings \cite{hawkes2017soft}. However, when a rigid cap is attached to the tip, this function is significantly compromised. Our soft caps ensure that this property is maintained while also allowing the robot to carry a payload at its tip.

\begin{figure}[h]
  \centering
  \includegraphics[width=1\linewidth]{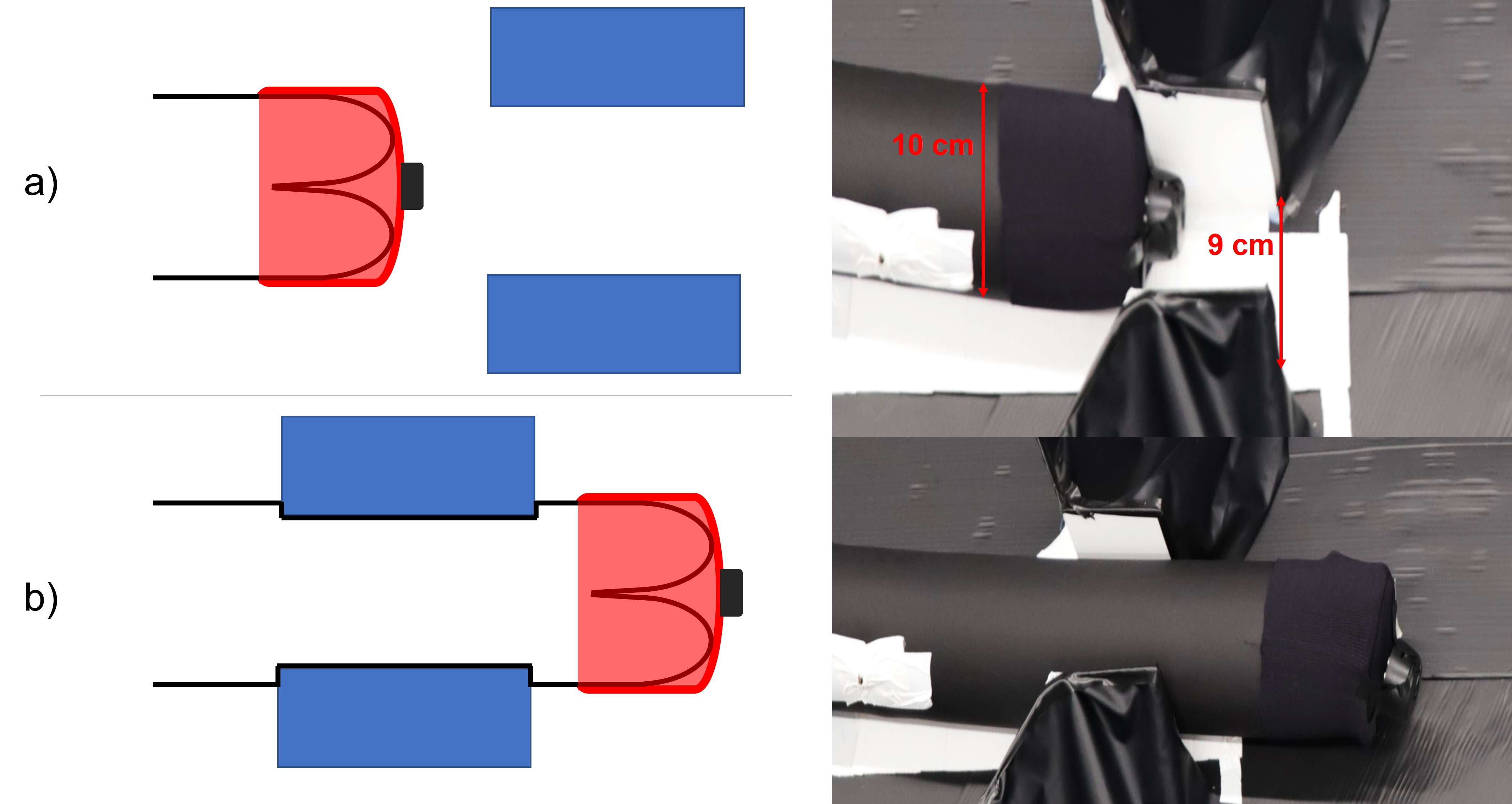}
  \caption{An eversion robot with a soft cap and camera attached to it, squeezing through a narrow opening. Diameter of the robot and width of the opening are indicated.}
  \label{figCh3}
\end{figure}

In this particular scenario, we constructed a gate with a width of 9 cm for a 10 cm diameter eversion robot. Using our soft cap, the robot was able to successfully pass through the narrow opening and continue its intended path while successfully delivering the payload (Fig. \ref{figCh3}).

\subsection{Quality 3: Navigability}

When a navigation actuation takes place, materials often expand, which limits the mobility of a rigid cap within a confined space. This problem can be avoided by using a soft cap, which can withstand great expansion without compromising tip mobility.


\begin{figure}[h]
  \centering
  \includegraphics[width=1\linewidth]{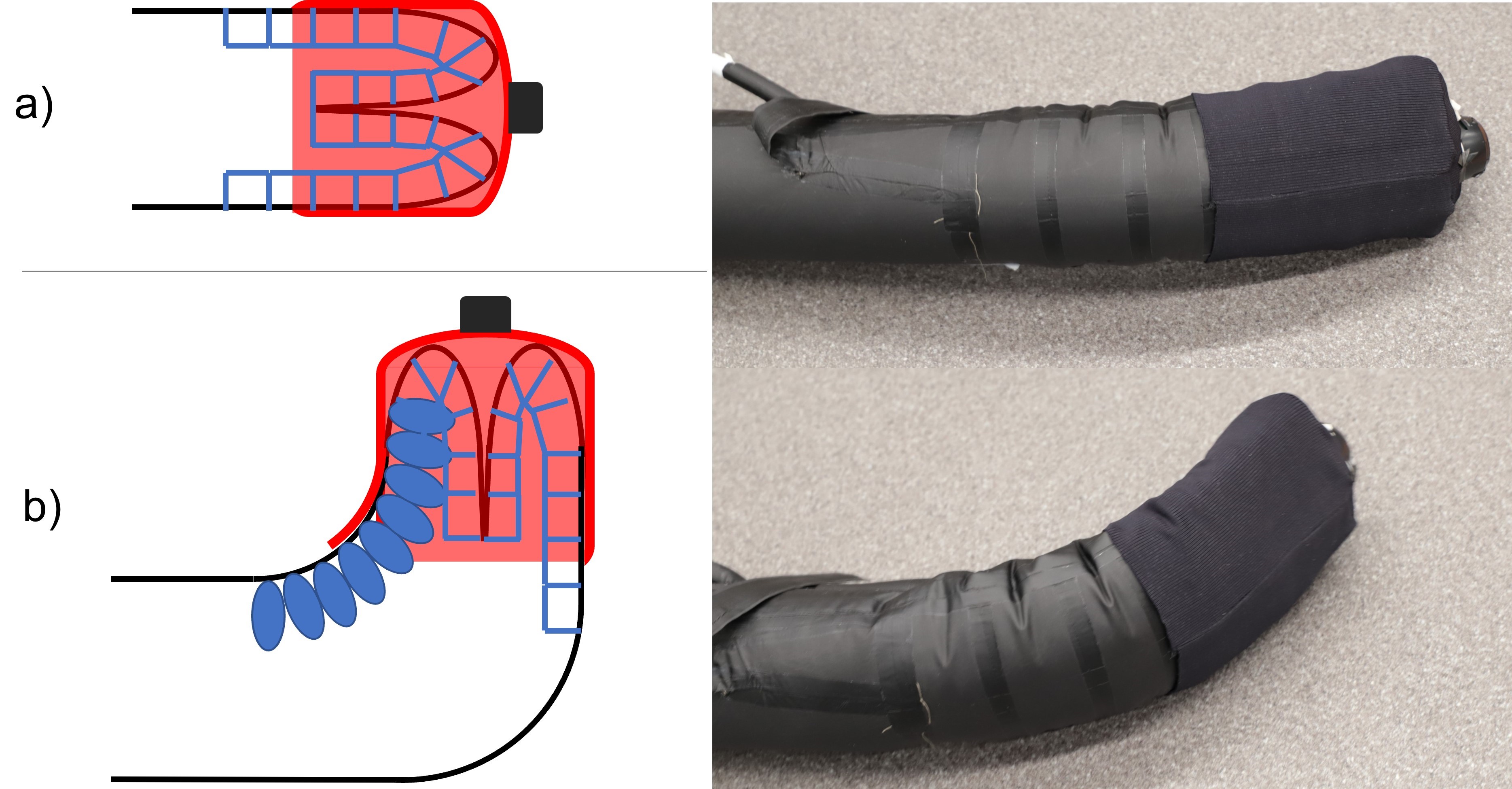}
  \caption{An eversion robot with a soft cap reorienting its tip.}
  \label{figCh4}
\end{figure}

In this scenario, we therefore activate our eversion robot with cap in situ, the challenge being to retain the cap at the tip while allowing requisite motion. At starting position the cap (Fig. \ref{figCh4} a) is attached to the eversion robot which is the halfway through eversion of the navigation layers (pouches \cite{abrar2021highly}), and then first the navigation layers then the main chamber of the robot are being inflated. The tip of the robot embraces the orientation change (Fig. \ref{figCh4} b) due to the navigation mechanism and keeps growing until it passes the navigation layers.

\section{Conclusions}

This extended abstract presents our caps for eversion robots made from elastic fabric parts. 
Exploiting textile properties, the cap adjusts to the everting robot and sits firmly, yet flexible at the tip of the robot's body throughout the eversion process. 
This design preserves squeezability and navigability of the robot, and the eversion of protruding elements is shown to be achievable. More available at \cite{softcap}.



\bibliographystyle{IEEEtran} 

\bibliography{references}
\vspace{12pt}

\end{document}